\documentclass[10pt, a4paper]{article}
\usepackage{lrec-coling2024} 

\usepackage{natbib}
\usepackage{multibib}
\makeatletter
\def\@mb@citenamelist{cite,citep,citet,citealp,citealt,citepalias,citetalias}
\makeatother
\newcites{languageresource}{~}

\usepackage{graphicx}
\usepackage{tabularx}
\usepackage{soul}
\usepackage{booktabs}
\usepackage{multirow}
\usepackage{multicol}
\usepackage{amsmath}
\usepackage{array}
\usepackage{graphicx}
\usepackage{titlesec}
\titleformat{\section}{\normalfont\large\bfseries\center}{\thesection.}{1em}{}
\titleformat{\subsection}{\normalfont\SmallTitleFont\bfseries\raggedright}{\thesubsection.}{1em}{}
\titleformat{\subsubsection}{\normalfont\normalsize\bfseries\raggedright}{\thesubsubsection.}{1em}{}
\renewcommand\thesection{\arabic{section}}
\renewcommand\thesubsection{\thesection.\arabic{subsection}}
\renewcommand\thesubsubsection{\thesubsection.\arabic{subsubsection}}

\usepackage{xcolor}
\usepackage{hyperref}
 \definecolor{darkblue}{rgb}{0, 0, 0.5}
  \hypersetup{colorlinks=true, citecolor=darkblue, linkcolor=darkblue, urlcolor=darkblue}

\usepackage{xstring}

\usepackage{color}

\title{ChatEL: Entity Linking with Chatbots}

\name {
    Yifan Ding,
    Qingkai Zeng, 
    Tim Weninger \\
}

\address{
Department of Computer Science \& Engineering \\ 
University of Notre Dame \\ 
Notre Dame, IN, USA \\
    \{yding4, qzeng, tweninge\}@nd.edu
}

\abstract{
Entity Linking (EL) is an essential and challenging task in natural language processing that seeks to link some text representing an entity within a document or sentence with its corresponding entry in a dictionary or knowledge base. Most existing approaches focus on creating elaborate contextual models that look for clues the words surrounding the entity-text to help solve the linking problem. Although these fine-tuned language models tend to work, they can be unwieldy, difficult to train, and do not transfer well to other domains. Fortunately, Large Language Models (LLMs) like GPT provide a highly-advanced solution to the problems inherent in EL models, but simply naive prompts to LLMs do not work well. In the present work, we define ChatEL, which is a three-step framework to prompt LLMs to return accurate results. Overall the ChatEL framework improves the average F1 performance across 10 datasets by more than 2\%. Finally, a thorough error analysis shows many instances with the ground truth labels were actually incorrect, and the labels predicted by ChatEL were actually correct. This indicates that the quantitative results presented in this paper may be a conservative estimate of the actual performance. All data and code are available as an open-source package on GitHub at \url{https://github.com/yifding/In_Context_EL}. 
 \\ \newline  }

\begin{document}

\maketitleabstract

\section{Introduction}

The introduction of Large Language Models (LLMs) has injected enormous excitement and anxiety into the world of natural language processing (NLP) and artificial intelligence (AI) generally. Although their text-generation and unstructured reasoning capabilities appear to be outstanding~\cite{Raffel-JMLR'20-T5, Radford-OpenAI_Blog'18-GPT, Radford-OpenAI_Blog'19-GPT2, Brown-NIPS'20-GPT3, Touvron-arXiv'23-Llama2}, their ability to produce structured output remains underdeveloped and relatively unexplored~\cite{Zhu-AAAI'22-JAKET, Dong-arXiv'23-Head-to-Tail}. The entity disambiguation task of Information Extraction (IE) seeks to link text fragments that represent some real-world entity with a structured list of that entity in, say, a knowledge base or dictionary. Once linked, the existing data and its relationships within the knowledge base could be used to assist in a number of downstream processes. This is the goal of the present work: to use LLMs to link the text fragments in the documents to structured knowledge. 

The merging of deep neural network models like the transformers used in LLM with the symbolic reasoning capabilities of extant knowledge bases has been dubbed by the United States Defense Advanced Research Projects Agency (DARPA) as the \textit{Third Wave of AI}~\cite{Darpa-Darpa'18-Next_AI} and as necessary for the advancement of problem solving and reasoning in AI systems~\cite{Brooks-AI'81-Symbolic, Hitzler-Book'22-Neuro}. For such a vision to become reality, it is critical that LLMs be linked to the symbolic entities that they reference.

\begin{figure}[ht]
    \centering
    \includegraphics[width=1.0\linewidth]{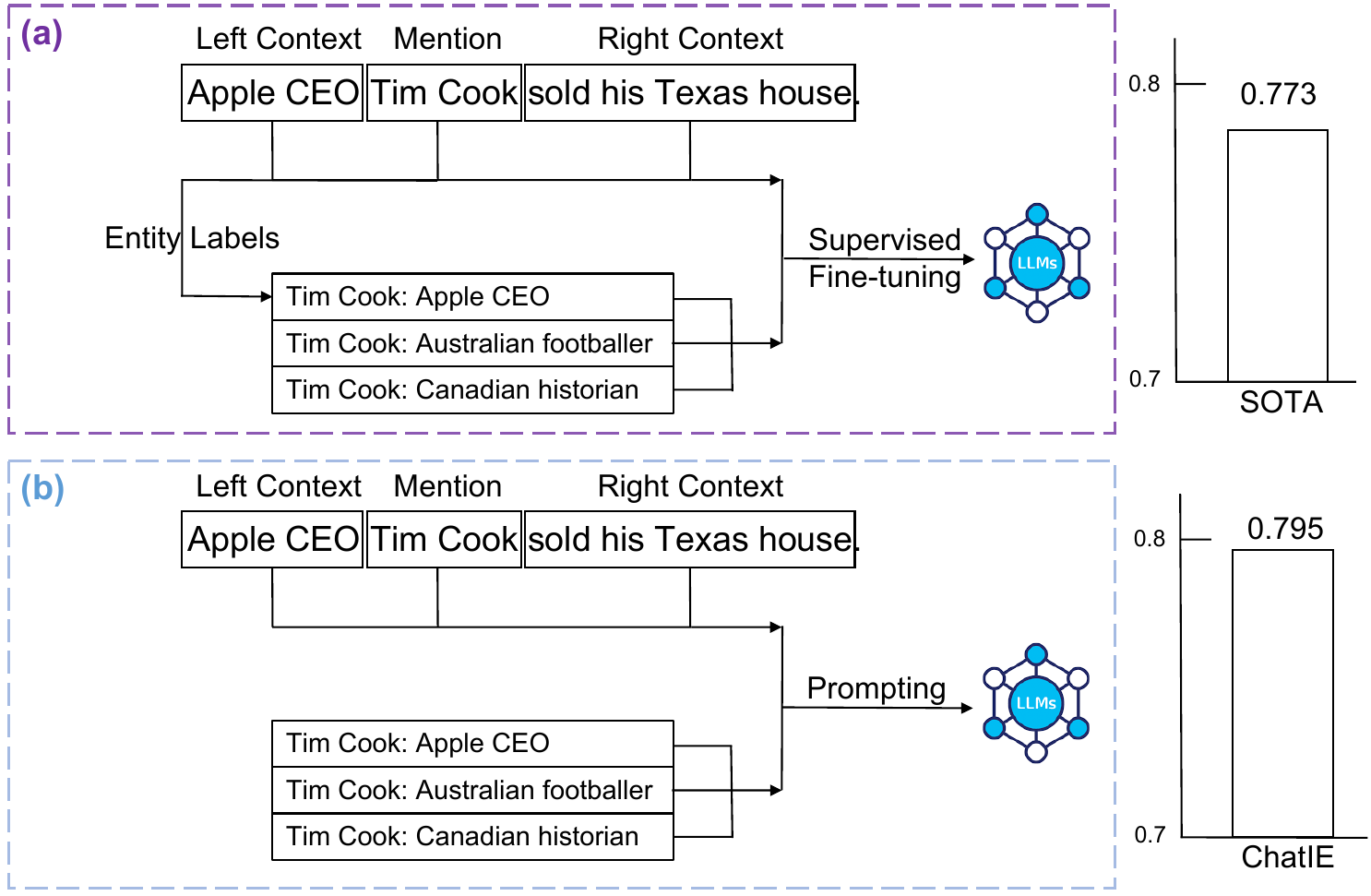}
    \caption{(a) General pipeline for supervised information extraction systems. These systems require careful modeling of the mention-text and its context and are fine-tuned on a large language model (LLM). (b) ChatEL relinquishes the context modeling entirely to the LLM and, instead, directly prompts LLM with the mention, context and entity candidates. ChatEL obtains a mean F1 score of 0.795 over ten datasets compared to 0.773 from the previous SOTA model. }
    \label{fig:intro}
    \vspace{-0.5cm}
\end{figure}

Perhaps unexpectedly, citation hallucinations~\cite{Emsley-Schizophrenia'23-ChatGpt_Hallucinations} within LLMs is a related problem. In these instances, the LLM generates a fake quote or citation from a fake source. Late-breaking work in citation-generation~\cite{Mohammad-ACL'09-Using_Citations} and text-grounding~\cite{Mun-CVPR'20-Text_Grounding} are working towards reconciling generated text with structured data. Likewise, recent work in prompt engineering and orchestration frameworks like LangChain~\cite{Chase-Softwar'22-LangChain} and Haystack~\cite{Pietsch_Haystack_the_end-to-end_2019} have begun to make inroads into generating structured output, like tables and lists. But in both cases, these systems do not perform information extraction because they do not link text snippets with external structured data. Recent research efforts~\cite{Qin-arXiv'23-Chatgpt_NLP_Task, Gonzalez-arXiv'23-ChatGPT_Historical_Entities} have utilized large generative models, specifically GPT-3.5 on information extracting tasks. However, these efforts are mainly focused on name entity recognition with tens of entity types, a far cry from the actual number (6 million) for general Wikipedia entities.

In the present work, we focus on the entity disambiguation task, one of the most difficult subtasks in IE, which requires ground text into actual entities. The state of the art in entity disambiguation is strongly rooted in supervised learning, where entity labels (Wikipedia pages or DBpedia entries) are predicted from a model trained on Wikipedia links~\cite{Ganea-EMNLP'17-deep_ed}, Web hyperlinks~\cite{Wu-EMNLP'20-BLINK}, or info-boxes~\cite{Ayoola-NAACL'22-ReFinED, Bhargav-NAACL'22-Dbpedia_Entity_Type}, considering inner-structure among entities~\cite{Hu-KBS'20-Graph_Entity}. 
And this has been shown to work in closed-world cases where the data is clean and fully available~\cite{Hoffart-EMNLP'11-CONLL03_entity} or in open-world scenarios where the data may be missing, but it is well-described~\cite{Logeswaran-ACL'19-zero, Wu-EMNLP'20-BLINK}. 

Another paradigm to deal with entity disambiguation task is through text generation. One of the earliest generative models to be used in entity disambiguation is called GENRE~\cite{DeCao-ICLR'21-GENDRE}. The GENRE proposed a sequence-to-sequence framework that could generate an entity-label sequence from a mention-sequence conditioned with some special indicators. However, like most existing entity disambiguation work, GENRE required full training from scratch, which required an enormous amount of data and hardware resources. 

However, supervised methods fail when the data is noisy, incomplete, poorly described, or rare. Recent work has found that at least 5\% of the ground truth labels on the entity-recognition task of the CONLL03 dataset are incorrect~\cite{Wang-EMNLP'19-Crossweigh, Chen-EMNLP'21-NLL, zeng2021validating}. Likewise, at least 10\% of the ground truth labels on the entity disambiguation task of the CONLL03 dataset are likely incorrect~\cite{Botzer-IPM'21-Reddit, Ding-NAACL_workshop'22-Posthoc}.

\paragraph{LLMs for Information Extraction}
Large language models like GPT-3.5~\cite{Ouyang-NIPS'22-InstructGPT}, GPT-4~\cite{Openai-arXiv'23-GPT4} and others have quickly supplanted individual modeling efforts that have permeated the NLP community for years. The incredible performance of large language models in zero-shot and few-shot settings has quickly replaced the previous pre-trained and fine-tuning approaches. How to design prompts for interaction with LLMs has become a highly regarded issue. The widespread availability of these prompt-based paradigms and their profound capabilities make them a natural ally in the improvement of many information extraction tasks. Recent efforts aim to prompt large pre-trained language models for various information extraction tasks, including named entity recognition and machine reading comprehension. The idea is to create some pre-defined template to query LLMs to obtain the desired output. Type-oriented prompts~\cite{Chen-ACL'23-MetaNER, Dong-arXiv'23-Head-to-Tail} aims to find corresponding mentions for desired class while span-oriented prompts~\cite{Li-arXiv'23-GPT-NER} aim to obtain the corresponding class for each span. Recently, PromptNER~\cite{Shen-ACL'23-Promptner} combined type-oriented methods and span-oriented methods to formulate NER into a finite set of ``ENTITY is TYPE'', and employed a dynamic template filling to assign the corresponding relationships. 

Unfortunately, adapting these frameworks to the entity disambiguation task requires some careful thought and experimentation. Entity disambiguation tasks are difficult primarily due to the unwieldy and generally undefined size of the class set. Specifically, encoding and discriminating millions of classes in entity disambiguation tasks with existing prompt methods are difficult and even unfeasible.

To solve this problem, we describe a straightforward and effective framework to utilize LLMs to assist in entity disambiguation task in this work. We propose a three-step framework based on prompting LLMs called ChatEL. ChatEL first generates a set of entity candidates for the mentions in the document. Then, it utilizes the power of the LLM to generate auxiliary content to support the selection of the corresponding entity from the candidates set.


We compared several state-of-the-art entity disambiguation models and evaluated them on ten public benchmarks. We show that ChatEL achieves comparable performance to the supervised models even without any training/fine-tuning on human-annotated data. We also conducted a further error analysis and showed two findings: 1) ChatEL was actually (sometimes) more reasonable than the answers provided on the ground truth; 2) In some cases, even ChatEL can not offer the correct prediction, the predictions of ChatEL are also highly-related to ground truth (\textit{e.g.}, hypernym of the ground truth entity). In short, our contributions can be summarized as follows:


\begin{itemize}
    \item We present ChatEL, a three-step Information Extraction tool that uses three-step prompts on LLMs. It successfully works on entity disambiguation task with millions of classes.
    \item We provide a comprehensive evaluation using ten datasets and compare the results with several state-of-the-art supervised models. Without supervised fine-tuning step, ChatEL matches and even outperforms supervised models with fine-tuning.
    \item We conduct a thorough error analysis. We show that the errant cases are oftentimes (arguably) more-correct than the ground truth. Also, when making mistakes, ChatEL still predicts quite close guesses on the ground truth.
\end{itemize}

\section{Related Work}
\subsection{Entity Disambiguation}
Entity disambiguation is one of the most challenging tasks in information extraction, as it aims to map annotated segments of a document to specific entities in a knowledge base. Existing research is primarily divided into two categories: improving feature extraction and refining task formulations. Key features include the interaction between mentions and context~\cite{Ganea-EMNLP'17-deep_ed,Kolitsas-CONLL'18-end2end}, consistency between entities and entity types~\cite{Tedeschi-EMNLP'21-NER4EL},relationship between entities and knowledge base entries ~\cite{Ayoola-NAACL'22-ReFinED}, and correlation between entities~\cite{Phan-TKDE'19-Pair_link}. 
Traditional entity disambiguation models framed the task as entity classification~\cite{Ganea-EMNLP'17-deep_ed} while more recent work approached it as machine reading comprehension~\cite{Barba-ACL'22-Extend}, dense retrieve~\cite{Wu-EMNLP'20-BLINK}, question answering~\cite{Zhang-ICLR'22-Entqa}, and sequence-to-sequence generation~\cite{DeCao-ICLR'21-GENDRE}.

\subsection{Prompt-based Learning for IE}
One of the pivotal developments in the field of prompt learning for information extraction was the release of GPT-3~\cite{Brown-NIPS'20-GPT3}. GPT-3 demonstrated remarkable capabilities in understanding and responding to natural language prompts. The benefit of prompt learning is the ability to query a language model to obtain its knowledge and understanding about a given context, showing significant ability in zero-shot and few-shot learning~\cite{Li-arXiv'23-GPT-NER, Dong-arXiv'23-Head-to-Tail}. Existing work on prompt learning for information extraction can be primarily categorized into two paradigms: type-oriented and span-oriented. Type-oriented methods~\cite{Ding-ACL'22-Prompt_Entity_Typing} mainly locate the given class for a certain mention within the original documents, while span-oriented methods~\cite{Cui-ACL'21-template_NER_BART} enumerate all possible spans and assign corresponding class labels.


\section{Problem Definition}

In this section, we present the key concepts used in this paper and formally define the entity disambiguation problem: Given a document represented as a sequence of tokens $\mathcal{D}$ and a set of subsequences $M = \{m_1, \cdots,m_n\}$ (\textit{i.e.}, mentions) containing in $\mathcal{D}$. The goal of the entity disambiguation task is to establish a mapping for each mention $m \in M$ to its corresponding entity $e$ in the set $\mathcal{E}$ representing entities in a knowledge base (such as Wikipedia or DBpedia).

\section{ChatEL: Information Extraction with Chatbots}

In the present work, we bring the power of LLMs to the information extraction task, especially entity disambiguation. The proposed ChatEL formulates entity disambiguation as a 3-step conditional selection: 1) Generate and filter entity candidates for LLM from knowledge base $\mathcal{E}$; 2) Augment each mention by extracting relevant information with a prompt for LLM; 3) Combine candidates from Step 1 with context from Step 2, forming a multi-choice question for LLM.

\subsection{Step 1: Entity Candidate Generation}

Given a mention $m$ in document $D$, we aim to find a subset of entity candidates $\mathcal{E}_c$ corresponding to $m$ within the KB $\mathcal{E}$. We combine two strategies to generate entity candidates. First, we deploy the Prior, which is based on the statistical information of hyperlinks~\cite{Ganea-EMNLP'17-deep_ed} and which allows us to obtain a subset of entity candidates $\mathcal{E}_p$ from $\mathcal{E}$ that exhibit syntactic similarity to the mentions within the document. However, the Prior suffers from low recall because it only considers syntactic similarity. In such cases, we also employ a dense retrieval model as a backup to augment the entity candidate generation. Specifically, we select the BLINK model~\cite{Wu-EMNLP'20-BLINK} as our retrieval model to generate extra entity candidates $\mathcal{E}_r$. The BLINK model constructs the dense retrieval model using cleaned Wikipedia hyperlinks. Therefore, the final entity candidates set from step 1 is $\mathcal{E}_c = \mathcal{E}_p \cup \mathcal{E}_r$, which includes 10 candidates.


\begin{figure}[t]
    \centering
    \includegraphics[width=0.9\linewidth]{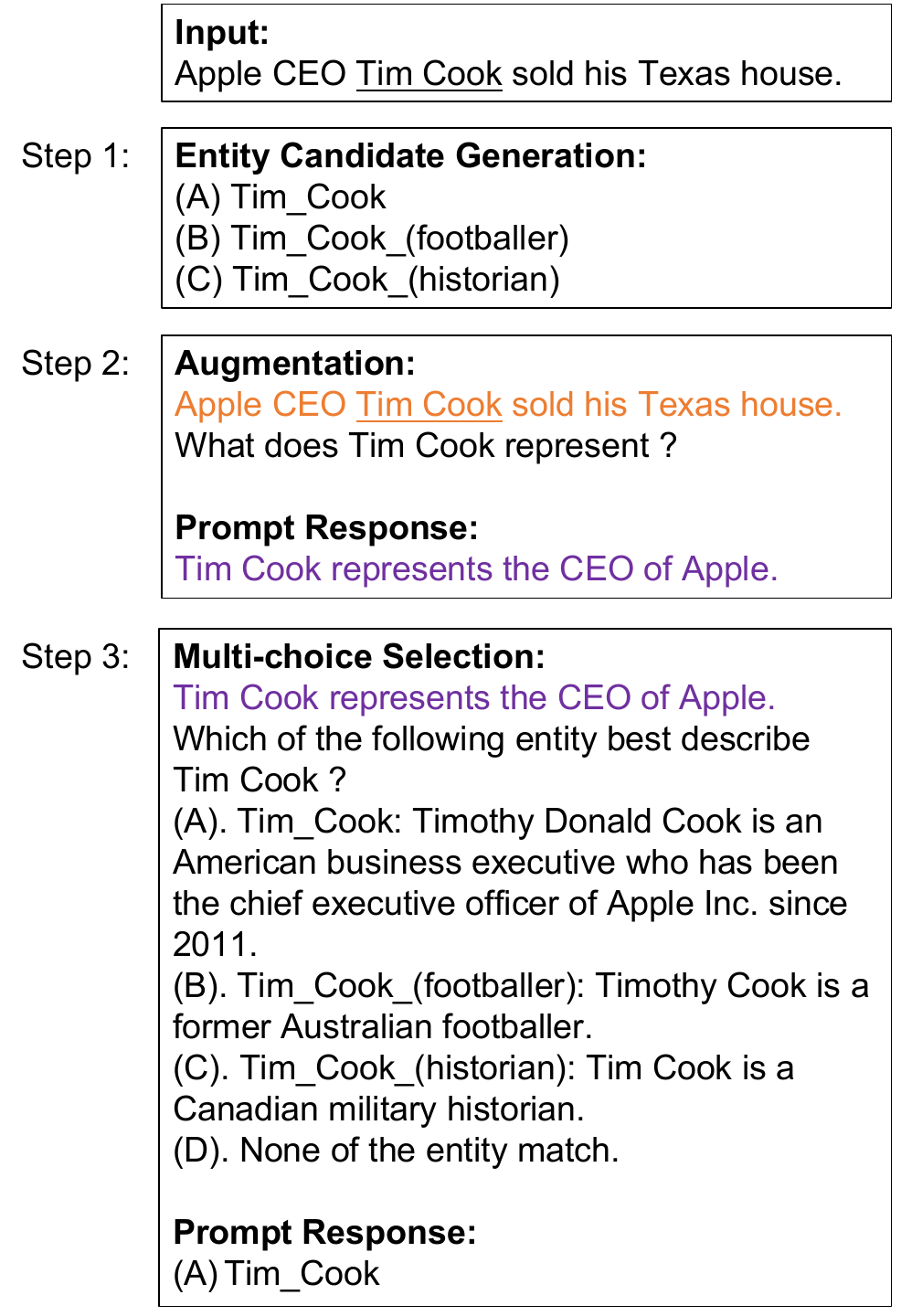}
    \caption{Pipeline of ChatEL framework: Given input document with the annotated mention, ChatEL first conducts (1) entity candidate generation step to obtain relevant entities. Then (2) an augmentation step is performed to obtain an auxiliary content of the annotated mention. Finally, (3) a multi-choice selection prompt is conducted to decide the corresponding entity of annotated mention.}
    \label{fig:model}
    \vspace{-0.5cm}
\end{figure}

\begin{table*}[ht]
\centering
\footnotesize
\caption{Data statistics for 10 experimental datasets. Number of documents (\# of Docs) and number of mentions (\# of Mention).}

\begin{tabular}{l | ccccccccccc}
\toprule 
 & \textbf{KORE} & \textbf{OKE15} & \textbf{OKE16} &
\textbf{REU} & \textbf{RSS} & \textbf{ACE04} &
\textbf{MSN} & \textbf{WIKI} & \textbf{AQU} & \textbf{CWEB}\\
\midrule
\# of Docs &50	&101	&173	&113	&357	&35	&20	&319 &50 &320 \\
\# of Mention &144	&536	&288	&650	&524	&257	&656	&6793	&727	&11154 \\

\bottomrule
\end{tabular}

\label{tab:dataset_stats}
\end{table*}

\subsection{Step 2: Augmentation by Prompting}
Since candidates are syntactically similar, distinguishing them is challenging without contextual information. To augment mentions with relevant information, we ask the LLM ``What does \textsf{Tim Cook} represent?'' 
(as shown in Fig.~\ref{fig:model}) to generate the auxiliary content $\mathcal{A}$ of ``\textsf{Tim Cook}'' based on the given document $\mathcal{D}$. We found that generating context information in this way has the following two advantages: 1) The content generated by the LLM is based on contextual information from the given document $\mathcal{D}$ and supplemented with the world knowledge encoded within the LLM. 

2) 
The auxiliary content produced by our system is more specifically targeted towards our task. Consequently, it significantly minimizes the impact of superfluous information.

\subsection{Step 3: Multiple-choice Selection by Prompting}

Given entity candidates set $\mathcal{E}_c$ (from step 1) and the auxiliary content $\mathcal{A}$ (from step 2), the goal of step 3 is to select the corresponding $e \in \mathcal{E}_c$. As shown in Fig.~\ref{fig:model}, we employ an instruct-based prompt to direct the LLM to make a selection from $\mathcal{E}_c$, utilizing the auxiliary content $\mathcal{A}$. To distinguish among these entity candidates effectively, we employ the first sentence extracted from the Wikipedia page of each entity candidate as a descriptive reference. It is noted that in step 3, all entity candidates come from the subset $\mathcal{E}_c$ obtained in step 1, rather than the complete KB $E$. Therefore, this multi-choice setting can not address the situation where the corresponding entity $e$ is not included in $\mathcal{E}_c$. To accommodate this situation, we include the option ``None of the entity match'' among the choices for handling such cases.

\section{Experiments}
Our proposed framework is evaluated on ten benchmarks. The experiments aim to address three research questions (RQs): 

\textbf{RQ1}: How does the performance of ChatEL compare to baselines in entity disambiguation?

\textbf{RQ2}: How do the components impact the performance of ChatEL?

\textbf{RQ3}: What are the reasons for ChatEL failing in some cases?

\subsection{Datasets}

Table~\ref{tab:dataset_stats} presents the statistics of ten benchmarks we used to evaluate the entity disambiguation task. In these ten benchmarks, there are five in-domain and five out-of-domain benchmarks each. All the experiments used Wikipedia as the background Knowledge Base (KB). Following the example of Guo and Barbosa~\citeyearpar{Guo-Semantic_Web'16-Random_Walk}, we manually removed spurious mentions that do not appear in the KB, as well as repeated documents, and empty documents without any mentions. The detail of these benchmarks are as follows:
\begin{itemize}
    \item \textbf{MSNBC (MSN), AQUAINT(AQU), ACE2004(ACE04):} All these benchmarks have annotated from news, aiming to link entities in the news with a KB~\cite{Cucerzan-EMNLP'07-MSNBC, Milne-CIKM'08-Learning_to_Link, Ratinov-ACL'11-Local_Global_Disambiguation}.
    \item \textbf{WNED-WIKI (WIKI), WNED-CWEB (CWEB):} These two large auto-extracted EL evaluation sets, were developed from ClueWeb and Wikipedia by ~\cite{Guo-Semantic_Web'16-Random_Walk, Gabrilovich-Dataset'13-Clueweb}. 
    \item \textbf{OKE-2015 (OKE15), OKE-2016 (OKE16):} They are from the Open Knowledge Extraction competition and are customized for ontology completion on DBpedia~\cite{Nuzzolese-SWEC'15-OKE15, Nuzzolese-SWEC'16-OKE16}. 
    \item \textbf{N3-Reuters-128 (REU), N3-RSS-500 (RSS):} N3-RSS-500 uses RSS feeds from global newspapers, covering various domains. N3-Reuters-128 includes economic news from Reuters-21587. Both datasets are manually annotated by~\cite{Roder-LREC'14-n3}.
    \item \textbf{KORE50 (KORE):} It contains brief, domain-varied documents from microblogging platforms like Twitter with ambiguous entity mentions~\cite{Hoffart-CIKM'12-KORE}.
\end{itemize}


\subsection{Baselines}
We compared the performance of ChatEL with the following baseline methods on entity disambiguation subtask:
\begin{itemize}
    \item \textbf{Prior}: It is a baseline string-matching algorithm that collects the entities corresponding to a given mention by looking through the entire Wikipedia corpus. We followed pre-processing in~\cite{Ganea-EMNLP'17-deep_ed}. The most frequent entity is selected as the target answer. 
    \item \textbf{REL~\cite{Hulst-SIGIR'20-REL}}: It is an entity disambiguation and entity linking package which combines the NER method FLAIR~\cite{Akbik-NAACL'19-FLAIR}.
    \item \textbf{End2End~\cite{Kolitsas-CONLL'18-end2end}}: It is a global entity disambiguation and entity linking system based on deep-ed~\cite{Ganea-EMNLP'17-deep_ed}, but with a more robust RNN architecture that considers mention-context and mention-mention interactions.
    \item \textbf{GENRE~\cite{DeCao-ICLR'21-GENDRE}}: This generation-based model considers entity disambiguation task as an entity name generation process.
    \item \textbf{ReFinED~\cite{Ayoola-NAACL'22-ReFinED}}: ReFinED is a recent system made by Amazon that considers Wikipedia entries as extra features to the entity disambiguation and entity linking tasks.
\end{itemize}

Since the performance of entity disambiguation highly relies on preprocessing strategies. In this work, we keep the original string from the dataset and consider all the non-empty entity as in-KB instance. During evaluation stage, we directly compare prediction entity string with the original entity string.


All data and code are available as an open-source package on GitHub at \url{https://github.com/yifding/In_Context_EL}.

\begin{table*}[ht]
\footnotesize
\caption{Test micro-F1 scores on 10 benchmarks. Best scores are highlighted in bold, second best scores are underlined. Gold-F1 is the upper bound performance of ChatEL, as all errors stem from ground truth entities not being included in the entity candidates set generated in step 1. All experiments are re-computed to compare entity names for evaluation\textsuperscript{1} }
{\renewcommand{\arraystretch}{1.2}

\begin{tabular}{l | ccccc | ccccc |c}
\toprule
\multirow{2}{*}{\textbf{Model}} & \multicolumn{5}{c|}{\textbf{Out-of-domain}} & \multicolumn{5}{c|}{\textbf{In-domain}} \\

& \textbf{KORE} & \textbf{OKE15} & \textbf{OKE16} &
\textbf{REU} & \textbf{RSS} & \textbf{ACE04} &
\textbf{MSN} & \textbf{WIKI} & \textbf{AQU} & \textbf{CWEB} & \textbf{AVG} \\

\midrule
\textbf{Prior} & 0.569	& 0.723	& 0.753	& 0.632	& \underline{0.756}	& 0.863	&0.903	&0.710	&0.864	&\underline{0.763}	& 0.754 \\
\textbf{REL} & \underline{0.618} & 0.705 & 0.749 & 0.662 & 0.680 & \textbf{0.897} & \textbf{0.930} & 0.783 & \textbf{0.881} & \textbf{0.771} & 0.768 \\
\textbf{End2End} & 0.569& \underline{0.767}& \underline{0.783}& 0.677& 0.720& 0.880& \underline{0.920}& 0.740& \underline{0.880}& 0.760& 0.770 \\

\textbf{GENRE} &0.542 &0.640 &0.708 &\underline{0.697} & 0.708 &0.848  &0.780 &\underline{0.823} &0.849 &0.659 &0.725 \\
\textbf{ReFinED} &0.567 &	\textbf{0.781} & \textbf{0.794}	& 0.680 &0.708	&0.864	&0.891	& \textbf{0.841}	&0.861	&0.738	&\underline{0.773} \\

\midrule
\textbf{ChatIE} & \textbf{0.787}	&0.758 & 0.752	& \textbf{0.789} & \textbf{0.822}	& \underline{0.893}	&0.881	&0.791 &0.767	&0.709  & \textbf{0.795}\\
\textbf{Gold-F1} &0.880	&0.903	&0.903	&0.911	&0.921	&0.969	&0.970	&0.944	&0.981	&0.943 & 0.932\\

\bottomrule
\end{tabular}

}
\textsuperscript{1}\scriptsize These performance reports may be different from the originally reported performance because of changes to the underlying datasets. Models tuned to out-of-date versions of the dataset may also have the names of the entries changed or removed resulting in performance degradation. 
\label{tab:main_experiment}
%
\end{table*}

\begin{table*}[t]
\centering
\footnotesize
\caption{Ablation study on 8 benchmarks (3 in-domain and 5 out-of-domain) with different backbone LLMs. The best scores are highlighted in bold, second best scores are underlined.
}
{\renewcommand{\arraystretch}{1.2}

\begin{tabular}{l| ccccc | ccc |c}
\toprule
\multirow{2}{*}{\textbf{Backbone}} & \multicolumn{5}{c|}{\textbf{Out-of-domain}} & \multicolumn{3}{c|}{\textbf{In-domain}} \\

& \textbf{KORE} & \textbf{OKE15} & \textbf{OKE16} &
\textbf{REU} & \textbf{RSS} & \textbf{ACE04} &
\textbf{MSN} & \textbf{AQU} & \textbf{AVG} \\

\midrule

\textbf{PaLM} &\underline{0.728}	&0.662	&0.665	&0.742	&0.767	&0.852	&0.814	&0.685	&0.739\\
\textbf{Llama-2-70B} &0.647	&0.617	&0.585	&0.649	&0.734	&0.746	&0.741	&0.635	&0.669\\
\textbf{GPT-3.5} & 0.716	& \textbf{0.767}	& \textbf{0.770}	& \underline{0.785}	& \underline{0.808}	& \textbf{0.918}	&\underline{0.867} & \textbf{0.791}	& \underline{0.803}\\
\textbf{GPT-4} & \textbf{0.787}	&\underline{0.758} & \underline{0.752}	& \textbf{0.789} & \textbf{0.822}	& \underline{0.893}	& \textbf{0.881} &\underline{0.767}	& \textbf{0.806} \\

\bottomrule
\end{tabular}

}
\label{tab:ablation_study_backbone}
\vspace{-0.2cm}
\end{table*}

\begin{table*}[ht]
\centering
\footnotesize
\caption{Ablation study on 8 datasets (3 in-domain and 5 out-domain) with GPT-4 backbone. The best scores are highlighted in bold. 
}
{\renewcommand{\arraystretch}{1.2}
\begin{tabular}{l | ccccc | ccc |c}
\toprule
\multirow{2}{*}{\textbf{Ablation}} & \multicolumn{5}{c|}{\textbf{Out-of-domain}} & \multicolumn{3}{c|}{\textbf{In-domain}} \\

& \textbf{KORE} & \textbf{OKE15} & \textbf{OKE16} &
\textbf{REU} & \textbf{RSS} & \textbf{ACE04} &
\textbf{MSN} & \textbf{AQU} & \textbf{AVG} \\

\midrule
\textbf{ChatIE w/o Aug. (step 2)} & 0.707	&0.696	&0.687	&0.688	&0.767	&0.853	&0.821 &0.753	& 0.747\\
\textbf{ChatIE w/o BLINK (step 1)} & 0.722	& \textbf{0.769}	&0.748	&0.676	&0.794	&0.890	&0.878	&\textbf{0.865}	& 0.793\\
\midrule

\textbf{ChatIE} & \textbf{0.787}	&0.758 & \textbf{0.752}	& \textbf{0.789} & \textbf{0.822}	& \textbf{0.893}	& \textbf{0.881} &0.767	& \textbf{0.806} \\
\bottomrule
\end{tabular}

}
\label{tab:ablation_study}
\vspace{-0.2cm}
\end{table*}

\subsection{Evaluation Metrics}
To maintain a fair comparison across datasets, we use the in-KB micro-F1 score as our evaluation metric following the example of Guo and Barbosa~\citeyear{Guo-Semantic_Web'16-Random_Walk}. Specifically, being \textit{in-KB} requires that ground truth mentions correspond to existing KB entries. Empty or invalid mentions are removed in the evaluation process. Micro-F1 score is as averaged per-mention. Although a model may predict non-entities, each mention will always have some corresponding ground truth entity.

\subsection{Main Results (RQ1)}



The results of performance comparison over ten entity disambiguation benchmarks are presented in Table~\ref{tab:main_experiment}. The Gold-F1 reported in Table~\ref{tab:main_experiment} shows the upper bound performance of ChatEL if the corresponding entity is included in the entity candidates set generated by step 1. The overall performance is around $90\%$ indicating that the candidates set generated by step 1 can cover most cases in all benchmarks.  
We have three main observations. First, we find that the ChatEL framework using GPT-4 outperformed the second-best method by an average micro-$F_1$ score of +$2.2\%$. Specifically, the ChatEL obtained the best performance on three out-of-domain datasets including KORE50, Reuters-128 and RSS-500 with absolute improvements in the F1 score of $16.9\%$, $9.2\%$, and $6.6\%$, respectively, demonstrating the effectiveness of our methods.

Second, we can observe that ChatEL performs better on the out-of-domain benchmarks than the in-domain benchmarks while REL and End2End show strong performances on in-domain benchmarks. The main reason for this is the backbone model (word2vec) used in REL and End2End is trained on the domain-related corpus such as Wikipedia corpus. We also observe that REL and End2End have a significant drop in the performance rankings of REL and End2end on the two out-of-domain benchmarks (REU and RSS).

Third, we can notice that ChatEL is the only method without supervised fine-tuning (SFT). This indicates that ChatEL is free from human-annotated data. Compared to the baseline methods that rely on SFT, while ChatEL may not outperform them on certain benchmarks, it demonstrates greater adaptability to different domains than SFT-based baselines.

\begin{table*}[ht]
\centering
\footnotesize
\caption{ Error Analysis of ChatEL. Data represents the absolute number of errors for each dataset and the type of the error.
}
\begin{tabular}{rl | ccccc | ccccc }
\toprule
&\multirow{2}{*}{\textbf{Error Type}} & \multicolumn{5}{c|}{\textbf{Out-of-domain}} & \multicolumn{5}{c}{\textbf{In-domain}} \\

& & \textbf{KORE} & \textbf{OKE15} & \textbf{OKE16} &
\textbf{REU} & \textbf{RSS} & \textbf{ACE04} &
\textbf{MSN} & \textbf{WIKI} & \textbf{AQU} & \textbf{CWEB} \\

\midrule
\multirow{2}{*}{\textbf{FP}} & \textbf{Alternative Entity} &4	&26	&27	&34	&12	&11	&34	&507	&71	&1318 \\
& \textbf{Fail to Reject} &8	&17	&19	&47	&42	&7	&2	&402	&12	&440\\
\midrule
\multirow{2}{*}{\textbf{FN}} & \textbf{Miss GT} &8	&62	&63	&34	&32	&9	&38	&535	&125	&1558\\
& \textbf{Miss Candidate} &23	&78	&76	&59	&35	&8	&37	&315	&15	&753\\

\bottomrule
\end{tabular}

\label{tab:error_type}
%
\end{table*}

\subsection{Ablation Study (RQ2)}
To better understand the effects of different components, we conduct an ablation study across various aspects of the ChatEL. Due to budget constraints, we removed the two very large datasets WIKI and CWEB from the ablation study.
\subsubsection{Backbone LLMs for ChatEL}
We applied ChatEL on four backbone LLMs including GPT-3.5~\cite{Ouyang-NIPS'22-InstructGPT}, GPT-4~\cite{Openai-arXiv'23-GPT4}, PaLM~\cite{Chowdhery-arXiv'22-Palm}, and LlaMa-2~\cite{Touvron-arXiv'23-Llama2}. We have several observations as follows. First, both GPT-3.5 and GPT-4 consistently achieve top-tier performance across all benchmarks, ranking first and second, respectively. This indicates that language models with more parameters can significantly enhance the performance of ChatEL. Second, we found that GPT-3.5 performed similarly to GPT-4 backbone with only a small ($0.3\%$) decrease in the average $F_1$ score. Thus, ChatEL has the opportunity to attain equivalent performance at a reduced expense. Last, three out-of-domain datasets (\textit{i.e.}, KORE-50, Reuters-128, and RSS-500) found the best performance improvement with GPT-4 over GPT-3.5 ($+7.1\%$, $+0.4\%$, and $+1.4\%$ respectively). This observation means that GPT-4 has a stronger domain adaptation capability compared to GPT-3.5.




\subsubsection{Effectiveness of Step 1 and Step 2}

We conducted an ablation study on the eight benchmarks to verify the effectiveness of the entity candidates generation strategy (step 1) and augmentation by auxiliary content step (step 2). For step 1, we create variants of ChatEL by removing the entity candidates generated by BLINK.
We also tested selecting the corresponding entity without auxiliary content for the LLM.

As shown in Table~\ref{tab:ablation_study}, in experiments in all benchmarks, removing the auxiliary content harms the performance of ChatEL. This proves that auxiliary content enhances the connections between the mention and the target entity. Note that removing the entity candidates generated by BLINK hurts the performance of ChatEL on six benchmarks. That indicates that BLINK can improve the coverage of the entity candidates set. We also have observed that ChatEL performs better without BLINK candidates on OKE15 and AQU. This is because BLINK may introduce noise into the entity candidate set, which ultimately hinders ChatEL's performance.





%
\section{Arguing with the Teacher (RQ3)}
\label{sec:arguing}
Most studies on IE and NLP tasks mainly use quantitative analysis for model performance evaluation, often overlooking errors in ground truth~\cite{Wang-EMNLP'19-Crossweigh, Chen-EMNLP'21-NLL}.. Recent analyses show a minimum of 5\% error rates in benchmarks like AIDA-CONLL~\cite{Ding-NAACL_workshop'22-Posthoc, Botzer-IPM'21-Reddit}. Such error-prone datasets are commonly used in entity disambiguation and linking research, questioning whether mismatches in results are actual errors from the model or issues stemming from the dataset's own inaccuracies.



\begin{figure}[t!]
    \centering
\includegraphics[width=0.9\linewidth]{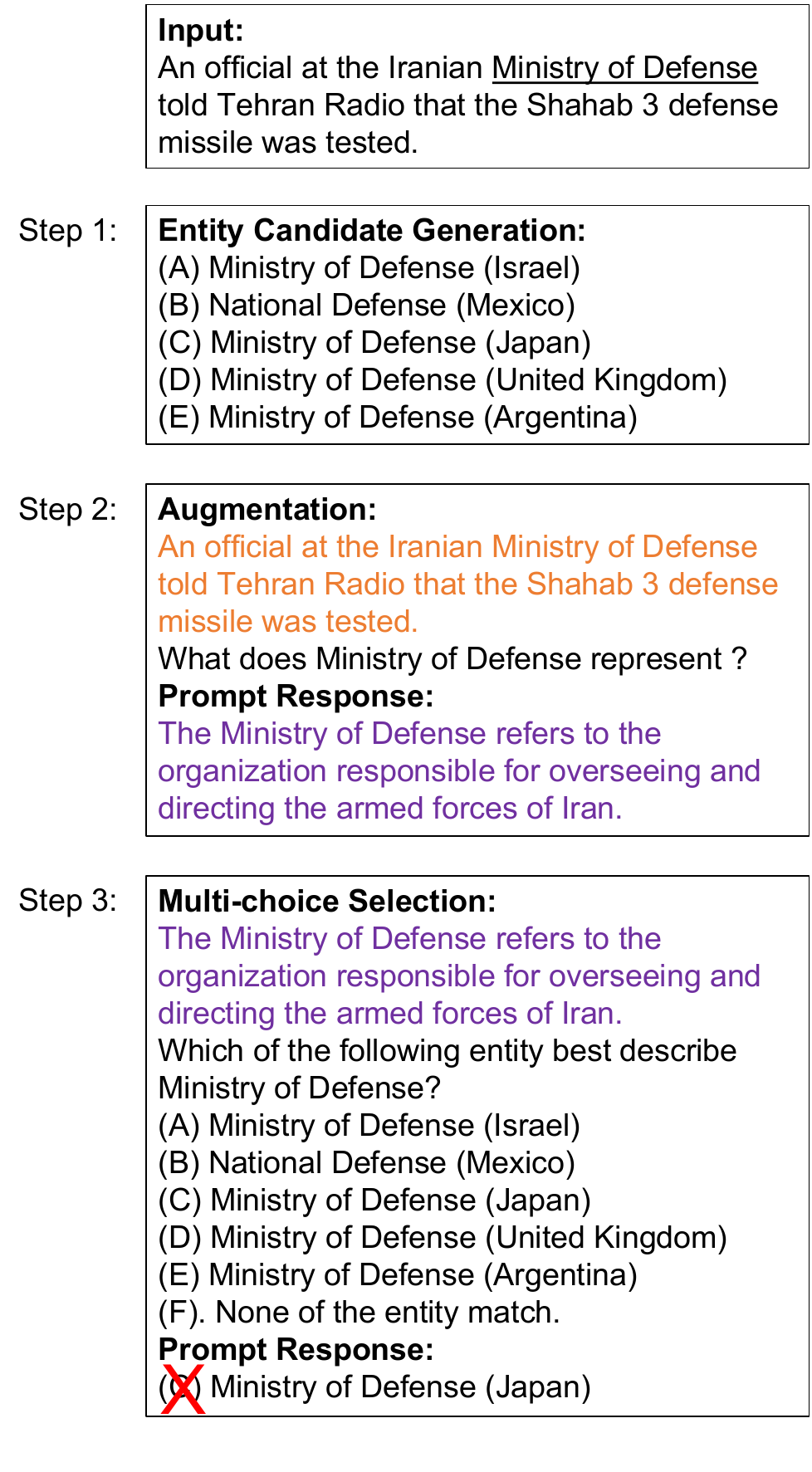}
\vspace{-0.5cm}
    \caption{Error case of ChatEL predicting Ministry of Defense (Iran) vs Ministry of Defense (Japan).}
    \label{fig:error_case}
    \vspace{-0.5cm}
\end{figure}

\begin{table*}[t]
    \centering
    \caption{Comparison of the ground truth label and the label predicted by the ChatEL model. We find that in many error cases the ChatEL model actually produces labels that are more accurate than the ground truth, and in several cases where the ground truth is indeed correct, the errant prediction is not far off.}
\scriptsize{
    \begin{tabular}{p{1cm}p{1.2cm}p{4.0cm}p{5cm}p{2.3cm}}
\toprule
\textbf{Dataset} & \textbf{Degree of Error} & \textbf{Ground Truth} & \textbf{Prediction} & \textbf{Reason} \\
\midrule
ACE04	&high	&Ministry of Defense (Iran)	&Ministry of Defense (Japan)	&Step 3  \\
ACE04	&low	&President of Egypt	&President	&Step 3 \\
ACE04	&low	&Gaza City	&Gaza Strip	&Step 2 \\
ACE04	&low	&Volvo	&Volvo Cars	&Step 3    \\
KORE	&low	&First Ladies of Argentina	&First Lady	&Step 2 \\
KORE	&high	&Justin Bieber	&Justin I	&Step 2 \\
KORE	&high	&Lady Gaga	&Gwen Stefani	&Step 3 \\
KORE	&high	&Paul Allen	&NULL	&Step 2 \\
AQU	&high	&Cancer	&Lung Cancer	&Step 3 \\
AQU	&high	&Tissue (biology)	&Facial tissue	&Step 3 \\
CWEB	&high	&Head	&Head (company)	&Step 2 \\
CWEB	&none	&Hillsborough County, Florida	&Hillsborough, North Carolina	&GT is incorrect \\
CWEB	&low	&Lake Wylie	&Lake Wylie, South Carolina	&Step 3 \\
CWEB	&none	&Australia Cricket Team	&Australia	&GT is incorrect \\
MSN	&none	&New York City	&New York	&GT is incorrect \\
MSN	&none	&University of Alabama	&Alabama Crimson Tide football	&GT is incorrect \\
MSN	&high	&World Trade Center	&Collapse of the World Trade Center	&Step 3 \\
OKE15	&none	&Fellow	&Research Fellow	&GT is incorrect \\
OKE15	&low	&Cambridge	&University of Cambridge	&Step 2 \\
OKE15	&none	&Principal (academia)	&Head teacher	&GT is incorrect \\
OKE15	&low	&Faculty (academic staff)	&Professor	&Step 2 \\
OKE15	&none	&Officer	&Officer (armed forces)	&GT is incorrect \\
OKE16	&none	&Director (business)	&Executive director	&GT is incorrect \\
OKE16	&none	&Germany	&Nazi Germany	&GT is incorrect \\
OKE16	&none	&Czechs	&Czech Republic	&GT is incorrect \\
OKE16	&none	&Sorbonne	&University of Paris	&GT is incorrect \\
REU	&none	&Georgia Power	&Georgia (U.S. state)	&GT is incorrect \\
REU	&low	&Lloyds Bank of Canada	&Lloyds Bank	&Step 3 \\
RSS	&none	&Steve Jobs	&Apple Inc.	&GT is incorrect \\
RSS	&low	&Pro Bowl	&Super Bowl	&Step 3 \\
RSS	&none	&Eric Kearney	&Cincinnati	&GT is incorrect \\
RSS	&high	&Cleveland Browns	&Cleveland	&Step 3 \\
\bottomrule
\end{tabular}
}
    \label{tab:errors}
\end{table*}

\subsection{Error Analysis}

We first compare the model predictions with the labels in the ground truth. In a case-by-case investigation, we identified two sets of two types of errors each. In the first case, we encounter false positives where the model may miss the correct label that is present in the list of candidates (Alternative Entity) or select a label from the candidates when the ground truth label was missing and the correct answer would be to select nothing (Failure to Reject), as illustrated in Fig.~\ref{fig:error_case}. In the second case, we encounter false negatives where the model incorrectly predicted the absence of a label, but a label did indeed exist. In these cases, two options were possible: first the correct ground truth label could have been present from the candidate list, but the model predicted no label (Missed Ground Truth) or the correct ground truth label could have been missing from the candidate list and the model did not find it (Missed Candidate). In these four cases, the errors can be recast as conditional on the ground truth being present or not in the list of candidates from which to pick. 

A detailed error analysis with this breakdown is presented in Tab.~\ref{tab:error_type}. From this table, we have some interesting observations. First, across all benchmarks, there does not appear to be a consistent distribution of error types. In all five out-domain datasets, the false negative error (Missed Ground Truth and Missed Candidate) appears to be more likely than the false positive error, indicating that the model hesitates to make predictions in these cases. In addition, we find most errors occur when the candidate entities do not contain the ground truth entity (row 2 and 4 in Tab.~\ref{tab:error_type}); that is, most errors occur when the ground truth entity is not in the list from which the label is picked---a case where the blame rests on the candidate generation step, not the model itself.


\subsection{Case Study}
In this section, we dive into ChatEL's error predictions to discern when and how ChatEL makes mistakes. We find that many of the predictions that are mismatched with the ground truth are actually more correct than the ground truth itself. 

Table~\ref{tab:errors} includes some cases from benchmarks in which the prediction unmatched the ground truth. We have annotated all the error cases of ChatGPT on the KORE50 and ACE04 datasets. After being revised by human experts, we found that the F1 performance increased by 2.2\% and 2.8\% respectively. This demonstrates the impact of incorrect ground truth data on the actual evaluation.


We also analyze the case in which ChatGPT did make mistakes and found that many times it makes reasonable predictions. For example, when the ground truth is the President of Egypt, it predicts ``President'' with the same part of speech but a broader meaning.  Furthermore, we analyzed what went wrong and discovered that step 3 makes the most mistakes. Additionally, step 2 sometimes makes mistakes by failing to provide useful auxiliary content, which leads to false negative predictions.



\section{Discussion}

In this work, we propose ChatEL, a three-step framework that leverages prompts to provide context that LLMs can use to link entity mentions from free text to their corresponding entries in a knowledge base. Unlike previous frameworks that produce complicated models to properly contextualize mention-text, the ChatEL framework simply replaces that complicated context model with an LLM with outstanding results. Unlike existing state-of-the-art models, ChatEL does not require any fine-tuning and is more accurate on average.

Furthermore, the detailed analysis in Sec.~\ref{sec:arguing} appears to indicate that quantitative results presented in the results section may be overcounting false positives due to errors in the ground truth. As a result, we believe that the performance metrics presented in the present work are a conservative estimate of the actual performance of ChatEL.

\section*{Ethics Statement}

Code and analysis are publicly accessible on GitHub, ensuring reproducibility. Characterized as low-risk, it utilized publicly available datasets, curated from news and web sources, containing no personally identifiable information, and resistant to falsification or misuse for misleading/libelous info.

The work primarily impacts text generative models' reliability, improving them by linking to curated knowledge bases, addressing issues like hallucinations in LLMs and enhancing fine-grained information tasks.



\clearpage

\nocite{*}
\section{Bibliographical References}\label{reference}


\bibliographystyle{lrec-coling2024-natbib}
\bibliography{reference}

\end{document}